\documentclass[10pt,twocolumn,letterpaper]{article}
\usepackage[pagenumbers]{wacv} 

\definecolor{wacvblue}{rgb}{0.21,0.49,0.74}
\usepackage[breaklinks,colorlinks,allcolors=wacvblue]{hyperref}

\usepackage{multirow,multicol}
\usepackage{pifont}
\usepackage{booktabs,colortbl,xcolor}
\usepackage{diagbox}
\usepackage{array}

\newcolumntype{C}{w{c}{0.95cm}}
\newcommand{\dg}[1]{\cellcolor{gray!18}#1}

\newcolumntype{X}{w{c}{0.85cm}}

\title{What Limits Thermal Infrared Drone Detection? A Cross-Dataset Benchmark Analysis}

\author{
    Tamara R. Lenhard$^{1,2,3}$ 
    \and
    Jan Kels$^{1}$
    \and
    Tobias Koch$^{1}$\\\vspace{-0.4cm}\and
    {\normalsize$^{1}$Institute for the Protection of Terrestrial Infrastructures,
    German Aerospace Center (DLR), Sankt Augustin, Germany}\\
    {\normalsize$^{2}$ACIDA Lab, Technical University of Applied Sciences Würzburg-Schweinfurt, Schweinfurt, Germany}\\
    {\normalsize$^{3}$EUT+ Data Science Institute, European University of Technology, European Union}\\\vspace{-0.4cm}\and
    {\tt\small \{tamara.lenhard, jan.kels, tobias.koch\}@dlr.de} \vspace{-0.4cm}
}

\begin{document}
\maketitle

\begin{abstract}
Thermal infrared (IR) imaging enables drone detection with reduced dependence on ambient illumination, supporting continuous surveillance beyond the limitations of visible-spectrum sensing. However, substantial and often undocumented variability in camera systems, processing pipelines, and annotation practices across public IR drone datasets complicates the assessment of detector performance and cross-domain generalization. To examine these sources of variability and their implications for drone detection, we systematically characterize existing IR datasets and establish an artifact taxonomy reflecting their origins across the acquisition and processing pipeline, encompassing optical and sensor effects, tone mapping and image enhancement, and recording-related artifacts. Focusing on long-wave IR data, we investigate the impact of training data composition and model characteristics on YOLO-based detection performance, generalization, and condition-dependent failure modes. Our results demonstrate that detection performance is governed more strongly by training data composition than by YOLO generation or model capacity, while cross-dataset generalization exhibits pronounced asymmetries. Even the best-performing configuration fails to detect over 40\% of drones, with elevated miss rates for very small targets, cluttered urban backgrounds, and thermal crossover, highlighting the shortcomings of existing datasets.
\end{abstract}

\section{Introduction}
\label{sec:intro}
The growing accessibility and operational capabilities of \textit{unmanned aerial vehicles (UAVs)}, commonly known as drones, present increasingly complex challenges for low-altitude airspace surveillance and critical infrastructure protection~\cite{Schneider:2021}. Unauthorized or malicious drone operations pose risks to public safety, critical infrastructure, and operational continuity~\cite{Johnson:2026}, particularly in densely populated urban environments. Early and reliable drone detection is therefore essential for effective surveillance and threat mitigation~\cite{Johnson:2026}. Camera-based systems -- particularly those operating in the visible (RGB) spectrum -- offer a promising and cost-effective approach, combining passive operation and compatibility with existing surveillance infrastructure. Coupled with \textit{deep learning (DL)} architectures such as \textit{You Only Look Once (YOLO)}~\cite{Redmon:2016}, these systems support automated drone detection through real-time image analysis~\cite{Wang:2024,Lenhard:2025}. However, RGB-based detection remains sensitive to adverse illumination and visibility conditions~\cite{Johnson:2026}. By capturing emitted thermal radiation rather than reflected visible light, \textit{infrared (IR)} imaging reduces reliance on ambient illumination~\cite{Vollmer:2017}, providing a complementary modality for continuous, day-and-night drone surveillance.

While IR-based drone tracking has received considerable research attention, standalone frame-wise detection remains comparatively underexplored~\cite{Dong:2025}, despite its essential role in track initialization and target re-acquisition following tracking failures in operational surveillance systems~\cite{Zhu:2023,Yu:2023}. Although YOLO-based detectors are increasingly employed for IR drone detection, variations in datasets, training protocols, and evaluation methodologies complicate meaningful comparisons across studies~\cite{Dong:2025}. This challenge is compounded by substantial, often undocumented differences in camera systems, image processing, recording pipelines, and annotation practices across datasets (see Sec.~\ref{sec:related_work}). Consequently, the relative contributions of detector design and dataset characteristics to reported performance remain difficult to disentangle.

\begin{table*}[t!]
\centering\footnotesize
    \caption{\label{tab:datsets_overview}Overview of publicly available IR datasets for drone detection and tracking. Sequence and frame counts refer exclusively to IR data. Unreported IR-bands were inferred by visual inspection. \ding{51}/\ding{55} denote the presence/absence of on-screen display (OSD).\vspace{-0.1cm}}
    \begin{tabular}{|l|c|c|c|r|c|c|l|}
    \hline
        Dataset & Pub. Year & No. Seq. & No. Frames & Resolution & IR Band & OSD & Image Quality / Visual Artifacts \\
        & & & (k) & \multicolumn{1}{c|}{(px)} & & &\\
        \hline
        LRDDv3$^{\text{\ding{117}}}$~\cite{Peterson:2026} & 2026 & 38 & 29.6 &  640$\times$512 & LWIR & \ding{55} & shaprness, blur, pseudocolor mapping \\\hline
        
        CST Anti-UAV~\cite{Xie:2025} & 2025 & 220 & 238  & 640$\times$512 & \ \ \ LWIR$^{\text{\ding{72}}}$ & \ding{51} & sharpening \& compression artifacts \\\hline
        
        Anti-UAV410~\cite{Huang:2024} & 2023 & 410 & 438$^{\text{\ding{115}}}$  & 640$\times$512 & MWIR & \ding{51} & lens reflection, compression, motion blur \\\hline
        
        \multirow[c]{2}{*}{Hamburg2 (H2)~\cite{Bhattacharya:2023}}  & \multirow[c]{2}{*}{2023} & \multirow[c]{2}{*}{23} & \multirow[c]{2}{*}{65} & 640$\times$480 & \multirow[c]{2}{*}{LWIR} & \multirow[c]{2}{*}{\ding{55}} & compression artifacts, watermark, \\
        & & & & 1024$\times$768 &&& vignetting, motion blur, defocus blur\\\hline
        
        Anti-UAV600~\cite{Zhu:2023} & 2023 & 600 & 723$^{\text{\ding{115}}}$ & 640$\times$512 & MWIR$^{\text{\ding{72}}}$ & \ding{51} & defocus blur, compression artifacts\\\hline 
        
        IRDST-real~\cite{Sun:2023}  & 2023 & 1,760 & 40.65 & \multicolumn{1}{c|}{--}  & LWIR & \ding{55} & defocus blur, fixed pattern noise\\\hline 
        
        Anti-UAV~\cite{Jiang:2023} & 2021 & 318 & 297$^{\text{\ding{115}}}$ & 640$\times$512 & \ \ \ MWIR$^{\text{\ding{72}}}$ & \ding{51} &  vignetting, defocus blur, motion blur\\\hline
        
        Halmstad Data~\cite{Svanstroem:2021,Svanstroem:2021_Data} & 2021 & 365 & 109.5 & 320$\times$256 & LWIR & \ding{55} & compression artifacts, image noise\\\hline

        USC~\cite{Chen:2017} & 2017 & 8 & 36.6 & 320$\times$240 & LWIR$^{\text{\ding{72}}}$ & \ding{51} & BST halo \& interlacing artifacts\\\hline\noalign{\smallskip}
        
        \multicolumn{8}{l}{$^{\text{\ding{72}}}$ IR band inferred from visual inspection \hspace{0.3cm} $^{\text{\ding{115}}}$ number of boxes \hspace{0.3cm} $^{\text{\ding{117}}}$ available upon request\vspace{-0.2cm}}\\
 
    \end{tabular}
\end{table*}

To address this gap, this work makes three \textbf{contributions}: (i)~We systematically characterize publicly available IR drone datasets and establish a taxonomy of artifacts based on their origin along the imaging pipeline. (ii)~Focusing on \textit{long-wave infrared (LWIR)} imagery, the predominant spectral band in public datasets, we evaluate YOLOv8 and YOLO11 across training datasets, model architectures, and capacities. (iii)~We systematically examine detection failures in relation to target size, background characteristics, and and instance-level attributes to identify factors affecting detection reliability.

The remainder of the paper is organized as follows: Sec.~\ref{sec:related_work} reviews related work on IR-based drone detection and tracking and characterizes publicly available IR drone datasets. Sec.~\ref{sec:exp_setup} details the experimental setup, followed by the results in Sec.~\ref{sec:results}. Key findings, conclusions, and future directions are discussed in Sec.~\ref{sec:conclusion}.

\section{Related Work}
\label{sec:related_work}
This section provides a systematic overview of the state of the art in IR-based drone detection and tracking. It reviews existing methodological approaches, surveys publicly available datasets for model training and evaluation, and compares their key characteristics.

\subsection{Drone Detection \& Tracking} 
\label{sec:RW_drone_detection_and_tracking}
Within the IR domain, existing research has predominantly focused on tracking, largely driven by the Anti-UAV challenges~\cite{Zhao:2021,Zhao:2023,Huang:2024} and dedicated IR tracking benchmarks~\cite{Jiang:2023,Huang:2024}, while frame-wise IR drone detection as a standalone task remains comparatively underexplored~\cite{Dong:2025}. Tracking approaches can broadly be categorized as (i)~detection-based, associating frame-wise detections over time, (ii)~Siamese-based, matching an initialized target template against subsequent frames, and (iii)~hybrid, combining both principles~\cite{Dong:2025}. In practical tracking systems, frame-wise detection remains an essential component, supporting target initialization and re-detection after tracking failure~\cite{Zhu:2023,Yu:2023}.

Whether employed as standalone detectors or integrated into tracking pipelines, IR-based drone detection predominantly relies on YOLO architectures spanning multiple generations: YOLOv2~\cite{Svanstroem:2021}, YOLOv3~\cite{Huang:2026,Yu:2026}, YOLOv5~\cite{Bhattacharya:2023,Huang:2026,Yu:2026,Hao:2025}, YOLOv6~\cite{Bhattacharya:2023,Yu:2026}, YOLOv7~\cite{Bhattacharya:2023,Bhattacharya:2025,Yu:2026}, YOLOv8~\cite{Ustun:2025,Tavaris:2025,Yu:2026,Hao:2025}, YO\-LOv9~\cite{Yu:2026,Hao:2025}, YOLO11~\cite{Tavaris:2025,Peterson:2026,Yu:2026,Hao:2025}, and YOLO12~\cite{Yu:2026}. The adoption of both earlier and recent YOLO generations -- whether as primary detector or as comparison baseline -- suggests that model selection is not strictly tied to recency, with YOLOv5, YOLOv8, and YOLO11 being the most frequently employed variants. Beyond standard YOLO implementations, several studies introduce targeted architectural refinements to enhance detection performance~\cite{Bhattacharya:2025,Huang:2026} or computational efficiency~\cite{Yu:2026}. Less prevalent alternatives include Faster R-CNN~\cite{Chen:2017,Huang:2026} and RT-DETR~\cite{Ustun:2025,Huang:2026,Yu:2026,Hao:2025}. Other approaches leverage signal-processing techniques (\eg, filtering)~\cite{Kou:2023,Liu:2024} or integrate RGB and IR modalities to enhance detection robustness. However, this methodological breadth is accompanied by substantial heterogeneity in experimental design (\eg, dataset, training regimes, and evaluation protocols), thereby limiting direct cross-study comparability~\cite{Dong:2025}.

\subsection{IR Drone Detection \& Tracking Datasets} 
\label{subsec:existing_datasets}
This section provides a systematic characterization of public IR drone detection and tracking datasets (Tab.~\ref{tab:datsets_overview}). Beyond dataset size and spatial resolution, we analyze sensor technology and spectral band, in-camera processing, and the recording pipeline, with a particular focus on their influence on visual characteristics and artifact patterns. For consistency, \textit{camera system} denotes the sensor, optics, and on-board processing, whereas \textit{recording pipeline} comprises all subsequent acquisition and storage stages.

\begin{figure*}[t!]
\centering
  \includegraphics[width=0.89\textwidth, trim={0.1cm 19.5cm 0.1cm 0.1cm}, clip]{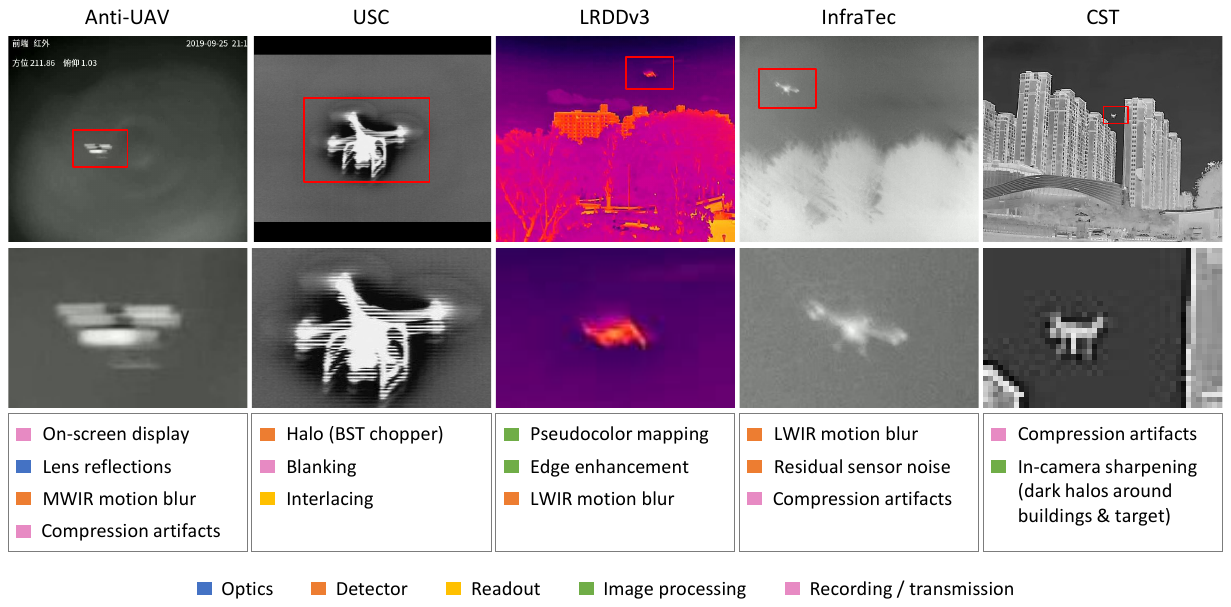}
  \caption{\label{fig:RW_Datasets_Artifacts} Selected examples of artifacts along the imaging pipeline in public IR drone detection and tracking datasets (left to right): Anti-UAV~\cite{Jiang:2023}, USC~\cite{Chen:2017}, LRDDv3~\cite{Peterson:2026}, InfraTec~\cite{Bhattacharya:2023}, and CST~\cite{Xie:2025}. \textbf{Top:} full-frame images with crop regions outlined in red (distinct from GT bounding boxes). \textbf{Bottom:} corresponding detail crops. Colors indicate stage of the imaging pipeline at which each artifact originates.}
\end{figure*}

\subsubsection{Dataset Scale}
Dataset scale is commonly reported in terms of sequence and IR-frame counts (Tab.~\ref{tab:datsets_overview}). Reported sequence counts vary substantially across datasets, although this variation partly reflects differences in recording and partitioning conventions. For instance, IRDST-real~\cite{Sun:2023} comprises a total of 1{,}760 sequences and 40.65k frames, whereas USC~\cite{Chen:2017} contains only 8 sequences but 36.6k frames. Individual sequence durations also differ considerably, ranging from 10\,s in Halmstad Data~\cite{Svanstroem:2021_Data} to several minutes in USC~\cite{Chen:2017}. Comparability is further affected by the subdivision of longer recordings into shorter clips, as in Anti-UAV~\cite{Jiang:2023}. Frame counts range from 29.6k in LRDDv3~\cite{Peterson:2026} to 238k in CST Anti-UAV~\cite{Xie:2025}. Depending on the dataset, individual frames can contain multiple drones, no drone, or other non-drone aerial objects (\eg, birds)~\cite{Svanstroem:2021_Data,Peterson:2026}. For tracking-specific benchmarks such as Anti-UAV~\cite{Jiang:2023}, Anti-UAV410~\cite{Huang:2024}, and Anti-UAV600~\cite{Zhu:2023}, dataset scale is even reported in annotated bounding boxes rather than frames. 

\subsubsection{Resolution \& Spectral Band}
Spatial resolution is primarily determined by the camera system. While earlier datasets such as USC~\cite{Chen:2017} and Halmstad Data~\cite{Svanstroem:2021_Data} employ 320-pixel formats, 640$\times$512 pixels has become the predominant resolution since Anti-UAV~\cite{Jiang:2023}. A notable exception is Hamburg2 (H2)~\cite{Bhattacharya:2023}, with subsets at 640$\times$480 and 1024$\times$768 pixels, the latter being the highest resolution among the reviewed datasets. 

Thermal camera systems for drone observation operate primarily in the mid-wave (MWIR, approx.\ 3-5\,\textmu m) or long-wave infrared (LWIR, approx.\ 7-14\,\textmu m)~\cite{Vollmer:2017}, as detailed in Tab.~\ref{tab:datsets_overview}. MWIR systems typically employ cooled photon detectors, whereas LWIR systems predominantly use uncooled microbolometers or, less commonly, pyroelectric \textit{barium strontium titanate (BST)} sensors~\cite{Vollmer:2017}. Beyond detector technology, MWIR and LWIR differ in their image-formation characteristics by sampling distinct regions of the electromagnetic spectrum. Under daylight conditions, MWIR imagery contains a substantial reflected-solar component, whereas LWIR imagery is governed predominantly by thermal emission~\cite{Handbook:1997}. Consequently, the spectral band influences the appearance and contrast of drones, backgrounds, and atmospheric effects. Despite its relevance, explicit spectral-band information is available for only five of the nine datasets listed Tab.~\ref{tab:datsets_overview}. For the remaining four datasets, the spectral band is inferred from sensor-specific imaging cues. Overall, six datasets were acquired in the LWIR and three in the MWIR band.

\subsubsection{Artifacts Along the Imaging Pipeline}
Thermal imagery across the considered datasets exhibits artifacts introduced at distinct stages of the imaging pipeline, spanning image formation, in-camera processing, and recording. Their stage of origin provides a natural taxonomy, separating camera-inherent degradations from downstream processing and recording artifacts.

\vspace{-0.3cm}
\paragraph{Optics \& Sensor.} Optical and sensor-level artifacts encompass both generic image degradations and detector-specific signatures. Common degradations include defocus~\cite{Jiang:2023,Bhattacharya:2023,Zhu:2023,Sun:2023}, vignetting~\cite{Jiang:2023,Bhattacharya:2023}, internal lens reflections~\cite{Huang:2024}, sensor noise~\cite{Svanstroem:2021_Data,Sun:2023}, and motion blur in both MWIR~\cite{Jiang:2023,Huang:2024} and LWIR datasets~\cite{Bhattacharya:2023,Peterson:2026}. Detector-specific physics further gives rise to characteristic imaging artifacts: in uncooled LWIR systems, the finite thermal time constant of microbolometers produces asymmetric motion trailing~\cite{Vollmer:2017}, as evident in LRDDv3~\cite{Peterson:2026}, whereas pyroelectric BST sensors require mechanical chopping, yielding characteristic halo artifacts in USC~\cite{Chen:2017}. Moreover, thermal cameras employ periodic shutter-based \textit{flat-field correction (FFC)} to compensate for sensor non-uniformity and thermal drift~\cite{Vollmer:2017}, typically causing brief acquisition interruptions followed by reduced fixed-pattern noise, as observed in CST Anti-UAV~\cite{Xie:2025}.

\vspace{-0.3cm}
\paragraph{Tone Mapping \& Image Enhancement.} The corrected 12-16\,bit signal from the sensor is typically mapped to an 8\,bit representation for visualization and storage via linear min-max normalization, as in H2~\cite{Bhattacharya:2023}, logarithmic remapping, or, most commonly, histogram equalization~\cite{Jiang:2023}. Under \textit{automatic gain control (AGC)}, the intensity mapping is adapted on a frame-by-frame basis to scene content, a behavior observed across the vast majority of datasets. Without knowledge of the applied dynamic range and transfer function -- which are typically not provided with the dataset -- this transformation is generally non-invertible, precluding recovery of the underlying sensor signal. Moreover, single-channel measurements are often mapped to pseudocolor representations, as in LRDD\-v3~\cite{Peterson:2026}. Some surveillance-oriented cameras additionally apply spatially non-uniform, content-adaptive processing, including local contrast enhancement and sharpening~\cite{Xie:2025,Peterson:2026}. Such nonlinear, spatially variant operations perturb the monotonic relationship between sensor response and recorded intensity, invalidating direct radiometric interpretation in terms of relative or absolute temperature~\cite{Vollmer:2017}. 

\vspace{-0.3cm}
\paragraph{Recording.} During recording, auxiliary information such as timestamps~\cite{Jiang:2023, Huang:2024, Zhu:2023, Xie:2025}, headings~\cite{Jiang:2023, Huang:2024, Zhu:2023}, reticles~\cite{Huang:2024, Zhu:2023}, and temperature scales~\cite{Xie:2025} is frequently embedded as \textit{on-screen display (OSD)} overlays, while some datasets additionally contain watermarks~\cite{Bhattacharya:2023, Svanstroem:2021_Data}. Furthermore, several datasets exhibit artifacts from spatial and temporal mismatches between the camera system and the recording pipeline. Spatial acquisition and format discrepancies introduce resampling, blanking, and interlacing artifacts, as observed in USC~\cite{Chen:2017} (Fig.~\ref{fig:RW_Datasets_Artifacts}) where files are obtained at 720 pixel resolution from a 320 pixel wide sensor. Temporal discrepancies typically arise from frame-rate differences: export-restricted thermal cameras often operate below 9\,Hz~\cite{Andrasi:2017}, requiring frame replication, when encoded at conventional video rates (\eg, 25\,Hz), as observed in CST Anti-UAV~\cite{Xie:2025}. Image and video compression further introduces coding artifacts across several datasets~\cite{Svanstroem:2021_Data,Bhattacharya:2023,Xie:2025}.

\subsubsection{Label Consistency}
Beyond file-format differences, the datasets exhibit substantial variation in bounding-box conventions. Box extents range from tight localization of the drone body~\cite{Xie:2025} to inclusion of arms and motion-blurred propellers~\cite{Bhattacharya:2023}, with some datasets employing more permissive bounding-box definitions (\eg, \cite{Peterson:2026}). Similar inconsistencies arise for motion-blurred, partially occluded, border-truncated, and low-visibility targets, which are retained in some datasets but excluded in others. Within individual datasets, we further observe annotation omissions and localization inaccuracies, including occasional boxes with little or no target overlap~\cite{Peterson:2026}. Such errors can compromise both training supervision and localization-based evaluation, particularly at stricter IoU thresholds.

\begin{table}[t]
\centering\footnotesize
\caption{\label{tab:YOLO_Specs}Capacity and computational complexity of evaluated YOLO configurations. Parameter counts and FLOPs refer to CO\-CO-pretrained models at an input resolution of 640$\times$640.\vspace{-0.1cm}} 
\begin{tabular}{|l|c|c|c|}
\hline
Model & Scale & Params (M) & FLOPs (B)\\\hline
\multirow{4}{*}{YOLOv8}
& n & 3.2  & 8.7   \\\cline{2-4}
& s & 11.2 & 28.6  \\\cline{2-4}
& m & 25.9 & 78.9  \\\cline{2-4}
& l & 43.7 & 165.2 \\\hline
\multirow{3}{*}{YOLO11}
& s & 9.4  & 21.5 \\\cline{2-4}
& m & 20.1 & 68.0 \\\cline{2-4}
& l & 25.3 & 86.9 \\\hline
\end{tabular}
\end{table}

\subsubsection{Summary \& Implications} 
All reviewed datasets exhibit artifacts arising from the choices throughout the acquisition and processing pipeline. When documented, these choices enable an informed composition of training data that includes artifacts expected in the target system (\eg, microbolometer motion blur) while excluding those that are not (\eg, compression artifacts). However, the availability of such documentation remains limited. Processing discrepancies between training and deployment (\eg, differences in color mapping) introduce domain shifts, while dataset-specific artifacts (\eg, OSD overlays) may provide misleading cues that encourage shortcut learning~\cite{Geirhos:2020} and amplify dataset bias~\cite{Torralba:2011}. For instance, OSD overlays may adversely affect model predictions depending on their presence in training and deployment, while irreversibly obscuring image content and potentially occluding targets. 

\section{Experimental Setup}
\label{sec:exp_setup}
This section details the experimental setup, including the considered drone detectors and training protocol, datasets, evaluation metrics, and experimental configurations.

\subsection{Detectors \& Training Protocol}
\label{subsec:model_selection_plus_training}
Detector selection is intended to support a controlled assessment of architectural design, model capacity, and training-data composition. Given its prevalence in IR drone detection and broad adoption in the RGB domain, the YOLO family serves as a natural basis for the present analysis. More specifically, we consider YOLOv8~\cite{Varghese:2024} and YOLO11~\cite{Ultralytics:YOLOv11} to represent meaningful architectural developments rather than simply successive model generations. While YOLOv8 relies exclusively on convolutional feature extraction, YOLO11 incorporates incorporates position-sensitive self-attention (C2PSA) at the coarsest backbone stage, alongside revised convolutional blocks (C3k2)~\cite{Ultralytics:YOLOv11}. YOLOv8 is evaluated across four model scales (n, s, m, and l) to characterize capacity-dependent effects, while YOLO11 is evaluated at three corresponding scales (s, m, and l) to facilitate architectural comparisons at nominally matched model capacities (Tab.~\ref{tab:YOLO_Specs}). The x variants are excluded due to single-GPU memory constraints. All models are trained from scratch for single-class drone detection under a standardized training protocol. Training is conducted for 300 epochs without early stopping, using a batch size of 64 and an input resolution of 640$\times$640 pixels. All remaining hyperparameters follow the Ultralytics default configuration\footnote{Ultralytics YOLO v8.3.239 (commit \texttt{34ab454}).}, without architecture-specific tuning. Experiments are performed on a single NVIDIA Quadro RTX 8000.

\begin{table}[t!]
    \centering\footnotesize
    \caption{\label{tab:data_splits}Frames per dataset and split. \textit{FLIR} and \textit{InfraTec} denote disjoint H2 camera subsets. Test frames are sequence-disjoint from training and validation sets. "--'' indicates not applicable.\vspace{-0.2cm}}
    \begin{tabular}{|l|l|c|c|c|}
    \hline
    Dataset & Subsets & \multicolumn{3}{c|}{Split} \\
    & & train & val & test \\\hline
    
    \multirow{3}{*}{H2~\cite{Bhattacharya:2023}}
    & FLIR & 26,368 &  -- & 14,251\\\cline{2-5}
    & InfraTec & 23,373 & -- & 2,446\\\cline{2-5}
    & Total & 49,741 & -- & 16,697\\\hline
    
    LRDDv3~\cite{Peterson:2026} & \multicolumn{1}{c|}{--} & 18,460 & 3,379  & 7,623 \\ \hline
    
    CST~\cite{Xie:2025} & \multicolumn{1}{c|}{--} & 124,301 & 43,527 & 70,214 \\ \hline
    \end{tabular}

\end{table}

\subsection{Datasets}
\label{subsec:data_selection}
Training and evaluation are restricted to the LWIR spectrum, which predominates among publicly available IR datasets, thereby avoiding cross-spectral
domain shifts. Among the LWIR datasets, USC~\cite{Chen:2017} and Halmstad Da\-ta~\cite{Svanstroem:2021_Data} are excluded due to their limited spatial resolution, and IRDST-real~\cite{Sun:2023} due to its emphasis on general IR small-target detection rather than drone-specific detection. The resulting selection comprises Hamburg2~\cite{Bhattacharya:2023}, LRDDv3~\cite{Peterson:2026}, and CST Anti-UAV~\cite{Xie:2025}, covering ground-to-air, air-to-air, and mixed acquisition settings, with the following characteristics:

\begin{table}[t!]
    \centering\footnotesize
    \caption{\label{tab:background_distribution_splits}Frame counts per background category across the CST~\cite{Xie:2025} splits. Categories are defined on a per-sequence basis, with each frame assigned to exactly one category.\vspace{-0.1cm}}
    \begin{tabular}{|l|c|c|c|c|c|c|}
    \hline
    Split & \multicolumn{6}{c|}{Background Category} \\ 
    & building & mountain & sky & jungle & urban  & water \\
    \hline
    train   & 53,648 & 20,230 & 8,470 & 4,908 & 26,177 & 10,868 \\\hline
    val     & 13,974 & 9,659 & 2,231 & 1,197 & 11,670 & 4,796 \\\hline
    test    & 8,458  & 10,687 & 6,611 & 13,651 & 26,004 & 4,803 \\\hline
    \end{tabular}
\end{table}

\begin{figure}
    \centering
    \includegraphics[width=0.47\textwidth, trim={0.3cm 25.9cm 8.7cm 0.1cm}, clip]{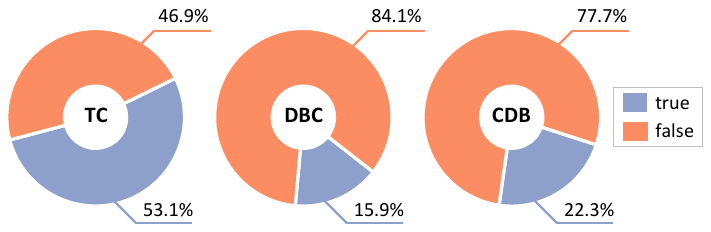}
    \caption{\label{fig:attrib_dist}Proportion of annotated drone instances exhibiting TC, DBC, and CDB across all splits of CST~\cite{Xie:2025}.}
\end{figure}

\begin{itemize}
    \item \textit{Hamburg2 (H2)}~\cite{Bhattacharya:2023} -- The H2 dataset comprises 66,438 sequential IR images with 71,520 annotated drone instances, acquired simultaneously with a FLIR Scion OTM366 (640$\times$480) and an InfraTec VarioCAM HD Z (1024$\times$768). The resulting scene-matched subsets (FLIR and InfraTec) differ in sensor, resolution, optics, and post-processing, enabling their use as distinct source and target domains. Recordings cover two distinct sites -- a university campus and a harbor -- with diverse backgrounds, drone models, and viewing distances. Approximately 14\% of the annotated instances are explicitly flagged as difficult due to poor target-background separability or substantial truncation at image boundaries.\vspace{0.1cm}
    
    \item \textit{LRDDv3}~\cite{Peterson:2026} -- LRDDv3 comprises 29,630 IR images at a spatial resolution of 640$\times$512 pixels. Recordings were captured at 30\,fps using an Autel Robotics EVO II Dual 640T V3 and subsampled to 5\,fps to increase background diversity and reduce temporal redundancy. Acquired in an air-to-air configuration, the dataset spans urban, rural, forested, and open-field environments, with target distances up to 200\,m. Birds and aircraft are annotated as distinct classes but treated as background during training.\vspace{0.1cm}
    
    \item \textit{CST Anti-UAV}~\cite{Xie:2025} -- The CST dataset comprises 220 IR sequences totaling $\sim$238,000 frames at a spatial resolution of 640$\times$512 pixels. The recordings were captured at 25\,fps using rotating platforms and UAV-mounted cameras. The dataset contains over 240,000 manually verified bounding-box annotations and covers a broad range of urban and natural environments under diverse seasonal, illumination (day-night, sunny-cloudy), and weather conditions (\eg, fog), including dynamic clutter and distractors (\eg, pedestrians, vehicles, birds, and aircraft). Additionally, CST provides frame-level attributes for \textit{occlusion (OCC)}, \textit{scale variation (SV)}, \textit{thermal crossover (TC)}, \textit{dynamic background clutter (DBC)}, and \textit{complex dynamic background (CDB)}. As OCC and SV (defined as changes in target size relative to the first frame) are primarily relevant to tracking, the analysis focuses on TC, DBC, and CDB, whose annotation frequencies are shown in Fig.~\ref{fig:attrib_dist}. Furthermore, each sequence and its constituent frames are assigned to one of six background categories: \textit{building}, \textit{mountain}, \textit{sky}, \textit{jungle}, \textit{urban}, or \textit{water} (Tab.~\ref{tab:background_distribution_splits}). While the training and validation splits exhibit similar background distributions, the test split shows substantial deviations.
\end{itemize}

\begin{figure}

    \includegraphics[width=0.44\textwidth, trim={0.2cm 24.9cm 10.3cm 0cm}, clip]{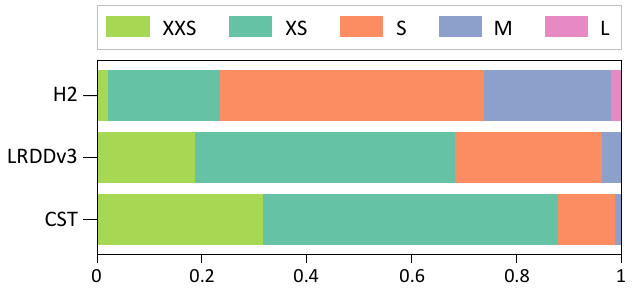}
    \caption{\label{fig:datasets_size_bucket_dist}Normalized object-size distributions for H2~\cite{Bhattacharya:2023}, LRDD\-v3~\cite{Peterson:2026}, and CST~\cite{Xie:2025}, showing annotation proportions by object-size category (Tab.~\ref{tab:size_categories}, Supp.) at 640$\times$640 YOLO input resolution.}
\end{figure}

\noindent All datasets provide predefined training and test splits, with additional validation splits available for CST and LRDDv3 (Tab.~\ref{tab:data_splits}). However, the partitioning strategies differ across datasets. H2 employs site-level separation, using harbor recordings for training and campus recordings for testing, whereas CST comprises 120/40/60 training/validation/test sequences, ensuring sequence-level separation. LRDDv3 reserves 13 video sequences exclusively for testing. Across all datasets, target sizes vary substantially, with small and extra-small instances predominating (Fig.~\ref{fig:datasets_size_bucket_dist}).

\subsection{Metrics}
\label{subsec:metrics}
Reliable drone detection requires minimizing both missed detections and false alarms to preserve operational effectiveness and system credibility~\cite{Lenhard:2024}. Accordingly, detection performance is characterized by the \textit{false negative rate (FNR)} and the \textit{false discovery rate (FDR)}, defined as $\mathrm{FNR}=\mathrm{FN}/(\mathrm{FN}+\mathrm{TP})$ and $\mathrm{FDR}=\mathrm{FP}/(\mathrm{FP}+\mathrm{TP})$. Here, $\mathrm{TP}$, $\mathrm{FN}$, and $\mathrm{FP}$ denote correctly detected targets, missed targets, and false detections, respectively. \textit{Mean average precision (mAP)} is additionally reported at \textit{intersection over union (IoU)} thresholds of 0.25 and 0.5, as well as averaged over thresholds from 0.5 to 0.95.

\subsection{Experiments}
\label{subsec:experiments}
Three experiments are conducted, progressing from training-data composition (Exp.~1) and detector design (Exp.~2) to a fine-grained analysis of failure modes (Exp.~3).

\vspace{-0.3cm}
\paragraph{Exp.~1 -- Cross-Dataset Generalization.} The first experiment evaluates detector generalization across datasets differing in camera systems, recording pipelines, and acquisition conditions. YOLOv8l is selected as the highest-capacity configuration considered to minimize the risk that dataset effects are masked by limited model capacity. Five YOLOv8l models are trained independently on CST, LRDDv3, the complete H2 dataset, and its FLIR and InfraTec subsets (Tab.~\ref{tab:data_splits}), following the standardized training protocol described in Sec.~\ref{subsec:model_selection_plus_training}. All models are evaluated across all five test splits using the metrics specified in Sec.~\ref{subsec:metrics}. For each evaluation metric, the resulting 5$\times$5 matrix distinguishes in-domain (diagonal) from cross-domain (off-diagonal) performance. As FLIR and InfraTec capture the same sites, targets, and flight events from different viewpoints and fields of view, their comparison enables a more controlled, albeit not fully matched assessment of camera-specific effects.

\vspace{-0.3cm}
\paragraph{Exp.~2 -- Model Benchmark \& Scaling Behavior.} The second experiment investigates the effects of detector architecture and model capacity on detection performance. All seven model configurations (Tab.~\ref{tab:YOLO_Specs}) are trained on the CST training split and evaluated on the corresponding test split using the metrics defined in Sec.~\ref{subsec:metrics}. CST is selected for its scale and diversity in environments, acquisition conditions, and recording platforms. Capacity-dependent effects are examined across model scales within each architecture, while architectural differences are assessed at nominally matched scales.

\vspace{-0.3cm}
\paragraph{Exp.~3 -- Failure Mode Analysis.}
Extending Exp.~2, the third experiment explores condition-dependent detection performance on the CST test split, focusing on three complementary dimensions: (i)~target size, (ii)~background category, and (iii)~instance attributes. The target-size analysis examines the sensitivity of detection performance to object scale, particularly given the limited spatial resolution of detector feature maps and the prevalence of small-scale targets (Fig.~\ref{fig:datasets_size_bucket_dist}). Background category is considered to assess the dependence of detection reliability on scene context, while instance attributes are examined to characterize performance variations associated with thermal crossover, dynamic background clutter, and complex dynamic background on the detection performance. For the target-size analysis, drone instances are categorized by GT bounding-box area according to an extension of the COCO size taxonomy~\cite{Lin:2014} to smaller targets (see Tab.~\ref{tab:size_categories}, Supp., for details). Detection performance is quantified separately for each size category using mAP and FNR. Background- and attribute-specific analyses leverage CST-exclusive metadata to evaluate detection performance across background categories and by attribute presence or absence, respectively. As background-based partitioning retains frames without target instances, FDR is evaluated exclusively for this dimension.

\section{Results}
\label{sec:results}
This section presents the experimental results outlined in Sec.~\ref{subsec:experiments}, focusing on cross-dataset generalization, the effects of detector architecture and model capacity, and condition-dependent detection performance.

\begin{table*}[t]
\centering\footnotesize
\setlength{\tabcolsep}{4.5pt}
\renewcommand{\arraystretch}{1.2}
\caption{\label{tab:results_v8l_cross-data_generalization}Cross-dataset generalization of YOLOv8l models trained on individual source datasets (rows) and evaluated on all test splits (columns). Shading indicates in-domain evaluations. Per-metric bests for each dataset are highlighted in \textbf{bold}. \vspace{-0.2cm}}
\begin{tabular}{|l| CCCCC @{\hspace{1.5em}} |CCCCC|}
\cline{2-11}
\multicolumn{1}{c|}{} & \multicolumn{5}{c|}{mAP@0.25 $\uparrow$}
& \multicolumn{5}{c|}{mAP@0.5-0.95 $\uparrow$} \\
\hline
\diagbox[width=5.8em,height=2.8em]{Train}{Eval} & CST & H2 & FLIR & InfraTec & LRDDv3 & CST & H2 & FLIR & InfraTec & LRDDv3 \\\hline
CST      
& \dg{\textbf{0.747}} & 0.732 & 0.751 & 0.543 & 0.010
& \dg{\textbf{0.452}} & 0.416 & 0.431 & 0.229 & 0.003 \\
H2   
& 0.291 & \dg{0.685} & 0.692 & 0.619 & 0.002
& 0.166 & \dg{\textbf{0.495}} & \textbf{0.504} & \textbf{0.404} & 0.000 \\
FLIR     
& 0.277 & 0.609 & \dg{0.625} & 0.459 & 0.001
& 0.121 & 0.431 & \dg{0.447} & 0.243 & 0.000 \\
InfraTec 
& 0.131 & 0.622 & 0.614 & \dg{\textbf{0.690}} & 0.095
& 0.065 & 0.401 & 0.405 & \dg{0.387} & 0.014 \\
LRDDv3   
& 0.426 & \textbf{0.746} & \textbf{0.764} & 0.619 & \dg{\textbf{0.418}}
& 0.088 & 0.358 & 0.371 & 0.222 & \dg{\textbf{0.105}} \\\hline

\multicolumn{11}{c}{\vspace{-0.3cm}}\\\cline{2-11}

\multicolumn{1}{c|}{} & \multicolumn{5}{c|}{FNR $\downarrow$} & \multicolumn{5}{c|}{FDR $\downarrow$}  \\\hline

\diagbox[width=5.8em,height=2.8em]{Train}{Eval}  
& CST & H2 & FLIR & InfraTec & LRDDv3
& CST & H2 & FLIR & InfraTec & LRDDv3 \\\hline

CST
& \dg{\textbf{0.537}} & \textbf{0.639} & \textbf{0.603} & 0.934 & 0.999 
& \dg{\textbf{0.030}} & 0.030 & 0.028 & 0.103 & 0.976\\

H2 
& 0.886 & \dg{0.762} & 0.751 & 0.854 & 1.000 
& 0.136 & \dg{\textbf{0.004}} & \textbf{0.003} & 0.019 & 0.000\\

FLIR     
& 0.834 & 0.775 & \dg{0.755} & 0.946 & 1.000 
& 0.178 & 0.059 & \dg{0.056} & 0.160 & 0.999\\

InfraTec 
& 0.864 & 0.822 & 0.828 & \dg{\textbf{0.768}} & 0.969 
& 0.760 & 0.011 & 0.012 & \dg{0.007} & 0.672 \\

LRDDv3   
& 0.879 & 0.660 & 0.627 & 0.936 & \dg{\textbf{0.783}} 
& 0.170 & 0.039 & 0.040 & \textbf{0.000} & \dg{\textbf{0.222}}\\
\hline
\end{tabular}
\end{table*}

\subsection{Cross-Dataset Generalization}
\label{subsec:generalization_results}
In-domain performance of YOLOv8l varies considerably across datasets (Tab.~\ref{tab:results_v8l_cross-data_generalization}). When trained and evaluated on CST, InfraTec, and H2, it achieves mAP@0.25 scores ranging from 0.685 to 0.747, compared with only 0.418 on LRDDv3. FNRs remain consistently elevated across all configurations (0.537-0.783 in-domain; $\geq$\,0.603 cross-domain), reflecting substantial target omission ($>$\,50\%) at the selected confidence threshold (0.25).

Under cross-domain evaluation, LRDDv3 and CST emerge as the most challenging target domains, with mean mAP@0.25 scores of 0.027 and 0.281, respectively, followed by InfraTec (0.560), H2 (0.677), and FLIR (0.705). Generalization also varies substantially with transfer direction: while YOLOv8l models trained on grayscale datasets perform poorly on LRDDv3 (mAP@0.25: $\leq$\,0.095), training on LRDDv3 yields considerably stronger cross-domain generalization, particularly on H2 and FLIR. Differences in thermal intensity representation between pseudocolor and grayscale imagery may contribute to this asymmetry, although variations in acquisition geometry and scene content preclude isolating this effect. A similar pattern can be observed for CST, which transfers relatively well to H2 and FLIR but remains challenging as evaluation target, with FDR rising from 0.030 in-domain to 0.760 cross-domain. Qualitative inspection further identifies recording and annotation artifacts that compromise even in-domain performance (Fig.~\ref{fig:detection_errors_artifacts}, Supp.): aircraft are consistently misclassified as drones, motion blur and compression artifacts cause intermittent misses, while OSD-occluded targets, inconsistent annotations of duplicated frames, and annotation misalignment undermine evaluation reliability. Despite its comparatively small training split (Tab.~\ref{tab:data_splits}), LRDDv3 achieves the strongest transfer to CST, suggesting that training volume alone does not determine cross-domain transferability.

On H2 and FLIR, CST- and LRDDv3-trained models even achieve higher mAP@0.25 scores than the in-domain baselines (CST: 0.732/0.751; LRDDv3: 0.746/0.764; in-domain: 0.685/0.625), with CST additionally yielding the lowest FNR. However, the in-domain baselines perform better at stricter IoU thresholds, partly reflecting their adaptation to dataset-specific bounding-box conventions. Recording artifacts further compromises target localization, as predicted bounding boxes incorporate adjacent OSD characters where overlays intersect the target.

The H2 subsets allow a more controlled assessment of camera-specific effects. While the InfraTec-trained model transfers to FLIR with minimal loss (mAP@0.25: 0.614~vs.~0.625 in-domain), the reverse transfer incurs substantial degradation (0.459~vs.~0.690), revealing pronounced domain shifts between camera systems. Temporal processing, particularly AGC, can further trigger abrupt detection failures, with targets remaining undetected after a contrast change (Fig.~\ref{fig:detection_errors_AGC}, Supp.).

\subsection{Model Benchmark \& Scaling Behavior}
\label{subsec:model_benchmark_results}
Performance is more sensitive to training data composition than to architectural choice or model capacity. On CST, mAP@0.25 ranges from 0.648 to 0.759 across model configurations (Tab.~\ref{tab:model_results_all_metrics}, Supp.), compared with 0.131 to 0.747 across training sources (Exp.~1). Increasing YOLOv8 capacity generally improves mAP and reduces FDR, although YOLOv8n exceeds YOLOv8s in mAP at all thresholds. The latter nonetheless achieves the lowest FNR (0.419) but highest FDR (0.367), reflecting an operating-threshold-dependent trade-off. YOLO11 exhibits limited performance scaling, with an mAP@0.25 gain of just 0.017 from s to l and s-m differences ($\leq$\,0.009) potentially attributable to run-to-run variability. At matched scales, YOLO11 achieves higher mAP@0.25 than YOLOv8 with fewer parameters, although the margin narrows from 0.094 at s to 0.012 at l. For instance, YOLO11s achieves comparable mAP@0.25 and lower FNR than YOLOv8l with just 22\% of the parameters and 13\% of the FLOPs.

\subsection{Failure Mode Analysis}
\label{subsec:failure_mode_analysis_results}
Failure modes are analyzed using YOLOv8l as reference model. Results for the remaining detectors, exhibit consistent trends and are provided in Sec.~\ref{subsec:SUPP_failure_mode_anaalysis_results} (Supp.). 

\begin{table}[t!]
    \centering\footnotesize
    \caption{\label{tab:results_YOLOb8l_by_Size_Cat}Performance of YOLOv8l on the CST test split by object-size category (defined in Tab.~\ref{tab:size_categories}, Supp.). Per-metric bests are shown in \textbf{bold}, while second-best results are \underline{underlined}.\vspace{-0.2cm}
    }
    \begin{tabular}{|l|ccc|c|}
    \hline
    Size & \multicolumn{3}{c|}{mAP~$\uparrow$} & FNR~$\downarrow$ \\
    Cat. & @0.25 & @0.5 & @0.5-0.95 & \\\hline
    XXS & 0.675 & 0.623 & 0.320 & 0.705 \\
    XS  & \underline{0.808} & \underline{0.799} & \underline{0.528} & \underline{0.452} \\
    S   & \textbf{0.848} & \textbf{0.842} & \textbf{0.569} & \textbf{0.343} \\
    M   & 0.553 & 0.543 & 0.391 & 0.690 \\
    L   & -- & -- & -- & -- \\    \hline
    \end{tabular}
    
\end{table}

\vspace{-0.3cm}
\paragraph{Target Size.} Detection performance depends strongly on target size (Tab.~\ref{tab:results_YOLOb8l_by_Size_Cat}). Small (S) targets are
detected most reliably, followed by XS targets, although S targets account for only 11.1\% of CST instances. Performance declines sharply for XXS targets, whose bounding-box dimensions fall below the finest YOLO feature-map stride, suggesting an inherent limitation in spatial resolution. Medium (M) targets also show reduced performance, likely reflecting their underrepresentation in CST ($\leq$\,1\% of annotations; Fig.~\ref{fig:datasets_size_bucket_dist}) rather than an inherent detection limitation. No large (L) targets occur in the CST test split.

\vspace{-0.3cm}
\paragraph{Background Category.} Detection performance is similarly influenced by background characteristics (Tab.~\ref{tab:results_YOLOb8l_by_BG_Cats}). More homogeneous backgrounds, such as \textit{water} and \textit{sky}, show particularly high mAP@0.25 values (0.999/0.986), although localization degrades at stricter IoU thresholds, particularly for \textit{sky}. In contrast, \textit{urban} and \textit{building} scenes exhibit the lowest detection performance, with the latter also showing the highest FDR (partly due to aircraft misclassified as drones; Fig.~\ref{fig:detection_errors_artifacts}, Supp.). Despite accounting for 43\% of training frames, \textit{building} scenes remain challenging, indicating that training prevalence alone does not ensure robustness in visually complex environments. Substantial variation also occurs within background categories. For instance, targets in \textit{mountain} sequences are detected more reliably against open sky than forested slopes (Fig.~\ref{fig:results_heatmaps_TP_FN}, Supp.).

\vspace{-0.3cm}
\paragraph{Instance Attributes.} All three CST attributes are associated with increased FNRs. The largest increase occurs for TC (from 0.377 to 0.653), where target temperatures converge with their surroundings, reducing thermal contrast. TC is also associated with a substantial decline in mAP and is the most prevalent attribute, affecting over half of all CST instances. CDB, characterized by multiple moving non-target objects, shows a slightly smaller increase (from 0.494 to 0.718), whereas DBC, reflecting dynamic changes in the target's immediate surroundings, has a comparatively minor impact. While these findings highlight the relevance of all three attributes, their individual effects cannot be fully disentangled from target and environmental characteristics.

\begin{table}[t!]
    \centering\footnotesize
    \caption{\label{tab:results_YOLOb8l_by_BG_Cats}YOLOv8l performance on the CST test split by sequence-level background category (trained on the CST training split). Per-metric bests are shown in \textbf{bold}; second bests are \underline{underlined}.\vspace{-0.2cm}}
    \begin{tabular}{|l|ccc|c|c|}
    \hline
    Background & \multicolumn{3}{c|}{mAP~$\uparrow$} & FNR~$\downarrow$ & FDR~$\downarrow$ \\
    Category & @0.25 & @0.5 & @0.5-@0.95 & & \\
    \hline
    water & \textbf{0.999} & \textbf{0.978} & \textbf{0.689} & \textbf{0.119} & \underline{0.013}  \\
    sky & \underline{0.986} & \underline{0.951} & 0.449 & \underline{0.222} & 0.033  \\
    jungle & 0.779 & 0.772 & 0.507 & 0.582 & \textbf{0.009}  \\
    mountains & 0.768 & 0.739 & \underline{0.513} & 0.511 & 0.020  \\
    urban areas & 0.617 & 0.602 & 0.364 & 0.678 & 0.024 \\
    building    & 0.573 & 0.565 & 0.325 & 0.643 & 0.111  \\\hline
    \end{tabular}
\end{table}

\begin{table}[t!]
    \centering\footnotesize
    \caption{\label{tab:results_YOLOb8l_by_Size_Attribute}YOLOv8l performance across CST test labels, stratified by the presence (\ding{51}) or absence (\ding{55}) of each annotated attribute.\vspace{-0.2cm}} 
    \begin{tabular}{|l|c|ccc|c|}
    \hline
    Attr. & Incl. & \multicolumn{3}{c|}{mAP~$\uparrow$} &  FNR~$\downarrow$ \\
     & & @0.25 & @0.5 & @0.5-0.95 &\\\hline
    \multirow[c]{2}{*}{TC}  & \ding{55} & \textbf{0.887} & \textbf{0.868} & \textbf{0.545} & \textbf{0.377} \\
                            & \ding{51} & 0.666 & 0.643 & 0.397 & 0.653 \\
    \hline
    \multirow[c]{2}{*}{DBC} & \ding{55} & \textbf{0.768} & \textbf{0.748} & \textbf{0.470} & \textbf{0.532} \\
                            & \ding{51} & 0.707 & 0.670 & 0.388 & 0.604\\
    \hline
    \multirow[c]{2}{*}{CDB} & \ding{55} & \textbf{0.792} & \textbf{0.770} & \textbf{0.481} & \textbf{0.494} \\
    & \ding{51} & 0.629 & 0.614 & 0.389 & 0.718 \\\hline
    \end{tabular}
\end{table}

\section{Discussion \& Conclusion}
\label{sec:conclusion}
The evaluation of publicly available LWIR drone datasets and YOLO-based detectors indicates that training data composition has a greater influence on detection performance than model architecture or capacity, with target size, background characteristics, and thermal crossover further affecting detection reliability. Both in-domain performance and cross-domain generalization remain constrained. Even the best configurations miss more than 40\% of all targets, a critical limitation for surveillance applications where missed detections pose significant operational risks. These limitations are partly attributable to the dataset characteristics and artifacts identified in Sec.~\ref{subsec:existing_datasets}: undocumented camera-specific processing introduces domain shifts and biases, recording artifacts such as OSD contribute to missed detections and false alarms, and annotation inconsistencies impair localization accuracy. Consequently, existing datasets alone appear to be insufficient to support reliable and transferable drone detection. Addressing these limitations requires datasets that systematically capture variations in target size, background, and thermal contrast, while ensuring transparent documentation of camera systems and recording pipelines, as well as consistent annotations.

Future work will explore synthetic data generation as a means of systematically addressing these requirements, drawing on the proposed artifact taxonomy to reproduce realistic degradations and examining the extent to which thermal rendering fidelity affects generalization of drone detectors to real-world surveillance scenarios.

\newpage
\twocolumn[{%
 \centering
 \vspace{1cm}
 \Large \textbf{Supplementary Material} \\[5em]
}]

\appendix
\setcounter{table}{0}
\setcounter{figure}{0}
\renewcommand{\thetable}{\Roman{table}}
\renewcommand{\thefigure}{\Roman{figure}}

\appendix

\section{COCO-based Object Size Categories}
\label{supp:size_categories}
Object size is quantified by the bounding-box area $A = w \times h$ in pixels (Tab.~\ref{tab:size_categories}). Under the COCO taxonomy~\cite{Lin:2014}, objects with $A < 32^2$ are classified as small, encompassing most drone instances in the evaluated datasets. To provide finer granularity, this category is further partitioned into xx-small, x-small, and small, following a scheme similar to~\cite{Hao:2025}. The xx-small threshold of $8^2$ pixels corresponds to the finest-resolution YOLO feature-map stride.

\begin{table}[h]
  \caption{\label{tab:size_categories} Object size categories by bounding-box area $A = w \times h$.\vspace{-0.2cm}}
  \centering\footnotesize
  \begin{tabular}{@{}|l|l|l|@{}}
    \hline
    Size Category & \multicolumn{1}{c|}{Object Area ($\text{pixels}^2$)} & Description\\\hline
    xx-small (XXS) & \ \ \ \ \ \ \ \  \ \ \ \ \ \ \ $A<8^2$ & extremely tiny objects \\\hline
    x-small (XS) & \ \ \ \ \ $8^2\leq A<16^2$ & very small objects\\\hline
    small  (S) & \ \ \ $16^2\leq A<32^2$ & small objects\\\hline
    medium (M) & \ \ \ $32^2\leq A<96^2$ & medium-sized objects\\\hline
    large  (L) & \ \ \ \ \ \ \ \  \ \ \ \ \ \ \ $A\geq96^2$ & large objects\\\hline
  \end{tabular}
\end{table}

\section{Extended Experimental Evaluation}
This section provides supplementary analyses complementing the results in Sec.~\ref{sec:results} (main paper), including additional evaluation metrics, complete model benchmarks, detailed failure mode analyses across detector configurations, and qualitative failure examples.

\begin{table}[t]
\centering\footnotesize
\setlength{\tabcolsep}{4.5pt}
\renewcommand{\arraystretch}{1.2}
\caption{\label{tab:results_v8l_cross-data_mAP0.25}Cross-dataset generalization of YOLOv8l at mAP@0.25. Rows and columns correspond to training and evaluation datasets, respectively. Shading indicates in-domain evaluations. The highest score per evaluation dataset is highlighted in \textbf{bold}.\vspace{-0.2cm}}
\begin{tabular}{|l|XXXXX|}
\cline{2-6}
\multicolumn{1}{c|}{} & \multicolumn{5}{c|}{mAP@0.5 $\uparrow$}\\\hline

\diagbox[width=5.8em,height=2.8em]{Train}{Eval} & CST & H2 & FLIR & InfraTec & LRDDv3\\\hline
CST & \dg{\textbf{0.726}} & \textbf{0.701} & \textbf{0.721} & 0.478 & 0.009 \\
H2 & 0.278 & \dg{0.679} & 0.689 & 0.585 & 0.000\\
FLIR & 0.234 & 0.600 & \dg{0.620} & 0.379 & 0.001\\
InfraTec & 0.119 & 0.616 & 0.608 & \dg{\textbf{0.685}} & 0.047\\
LRDDv3 & 0.288 & 0.689 & 0.713 & 0.416 & \dg{\textbf{0.284}}\\\hline
\end{tabular}
\end{table}

\begin{table}[t!]
    \centering\footnotesize
    \caption{\label{tab:model_results_all_metrics}Comparison of YOLO variants across model generations and architectural scales, trained on the CST training split and evaluated on the corresponding test split. Best results are highlighted in \textbf{bold}; second-best results are \underline{underlined}.\vspace{-0.2cm}} 
    \begin{tabular}{|l|ccc|c|c|}
    \hline
    Model & \multicolumn{3}{c|}{mAP~$\uparrow$} & FDR~$\downarrow$ & FNR~$\downarrow$\\
    & @0.25 & @0.5 & @0.5-0.95 & & \\\hline
    YOLOv8n & 0.657 & 0.637 & 0.386 & 0.084 & 0.538 \\
    YOLOv8s & 0.648 & 0.619 & 0.363 & 0.367 & \textbf{0.419} \\
    YOLOv8m & 0.711 & 0.691 & 0.420 & 0.062 & 0.532 \\
    YOLOv8l & \underline{0.747} & \underline{0.726} & \textbf{0.452} & \textbf{0.030} & 0.537 \\
    \hline
    YOLO11s & 0.742 & 0.717 & 0.434 & 0.068 & 0.472 \\
    YOLO11m & 0.734 & 0.708 & 0.431 & 0.064 & 0.496 \\ 
    YOLO11l & \textbf{0.759} & \textbf{0.731} & \underline{0.446} & \underline{0.047} & \underline{0.471} \\
    \hline
    \end{tabular}
\end{table}

\begin{figure*}[t!]
    \centering
    \includegraphics[width=\textwidth, trim={0cm 20cm 0cm 0cm}, clip]{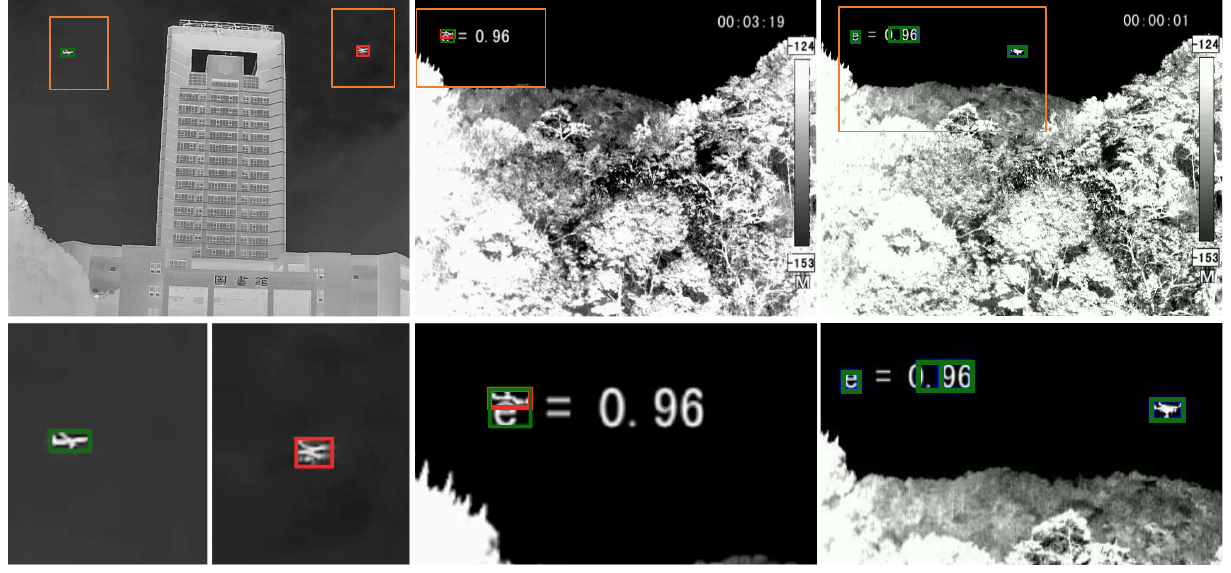}
    \caption{\label{fig:detection_errors_artifacts}Qualitative detection failures of YOLOv8l on the CST~\cite{Xie:2025} test split (red: GT; green: predictions; orange: regions enlarged below). \textbf{Left:} an aircraft is falsely detected as a drone, while the actual target is missed. \textbf{Middle:} an OSD character occludes the drone, shifting both the annotation and prediction onto the overlay. \textbf{Right:} OSD characters produce false detections alongside the correctly detected drone.}
\end{figure*}

\begin{figure*}[t!]
    \centering
    \includegraphics[width=\textwidth, trim={0cm 25.7cm 0cm 0cm}, clip]{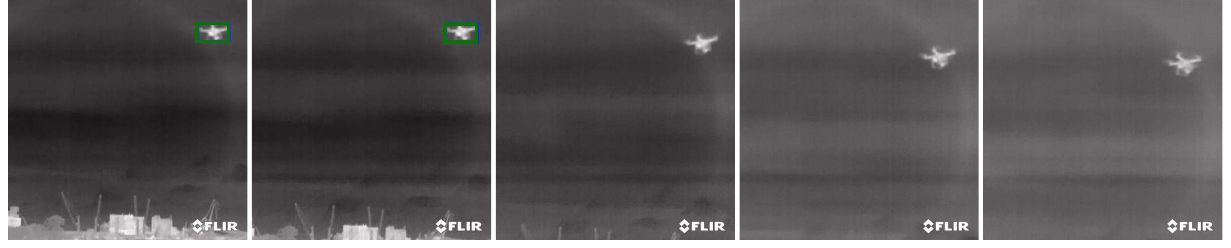}
    \caption{\label{fig:detection_errors_AGC}Consecutive frames from the H2 FLIR subset~\cite{Bhattacharya:2023}, illustrating the loss of drone detection (green box) following an apparent AGC-related contrast shift, despite the target remaining visible.}
\end{figure*}

\subsection{Cross-Dataset Generalization}
Further insights into cross-dataset generalization (Sec.~\ref{subsec:generalization_results}, main paper) are provided through mAP@0.5 evaluations and qualitative analyses of detection errors.

\vspace{-0.3cm}
\paragraph{Results at mAP@0.5.} The evaluation at mAP@0.5, reported in Tab.~\ref{tab:results_v8l_cross-data_mAP0.25}, broadly preserves the trends observed at mAP@0.25. On H2 and its FLIR subset, CST-trained models achieve the highest mAP@0.5 (0.701 and 0.721), surpassing the respective in-domain baselines (0.679 and 0.620). In contrast, LRDDv3-trained models, which lead at mAP@0.25, exhibit larger performance declines at the stricter IoU threshold (H2: 0.746 to 0.689; FLIR: 0.764 to 0.713; InfraTec: 0.619 to 0.416). This greater IoU sensitivity is consistent with LRDDv3's more permissive bounding-box conventions discussed in the main paper.

\vspace{-0.3cm}
\paragraph{Qualitative Failure Cases.} Qualitative analysis reveals recurring detection failures in both in-domain and cross-domain evaluations. On CST, YOLOv8l falsely identifies aircraft as drones while simultaneously missing the actual target (Fig.~\ref{fig:detection_errors_artifacts}, left). On-screen display (OSD) overlays introduce further ambiguities: overlapping characters distort target annotations and shift detections toward the overlay (middle), while isolated characters trigger false detections, suggesting reliance on spurious overlay features (right). Beyond recording artifacts, temporal in-camera processing introduces additional instability. In sequences captured by the FLIR camera in the H2 dataset, an abrupt global contrast shift, consistent with automatic gain control (AGC), coincides with the loss of a previously detected drone despite its continued visibility in subsequent frames (Fig.~\ref{fig:detection_errors_AGC}).

\subsection{Detector Performance across Model Scales}
Tab.~\ref{tab:model_results_all_metrics} provides the complete evaluation metrics for all seven detector configurations trained and evaluated on CST, complementing the summary in Sec.~\ref{subsec:model_benchmark_results} (main paper). YOLO11l achieves the highest mAP@0.25 and mAP@0.5, while YOLOv8l attains the highest mAP@0.5-0.95 and lowest FDR. YOLOv8s achieves the lowest FNR but the highest FDR, suggesting a more permissive operating point rather than superior overall detection performance. 

\subsection{Extended Failure Mode Analysis}
\label{subsec:SUPP_failure_mode_anaalysis_results}
Tabs.~\ref{tab:model_bucket_metrics}-\ref{tab:results_background_cats_all_models} extend the failure mode analysis in Sec.~\ref{subsec:failure_mode_analysis_results} (main paper) to the remaining detector configurations. Overall, the failure patterns identified for YOLOv8l remain evident across YOLO generations and model scales.

\vspace{-0.3cm}
\paragraph{Target Size.} Across configurations, S targets yield the highest mAP at IoU thresholds of 0.25 and 0.5 and the lowest FNR, followed by XS targets (Tab.~\ref{tab:model_bucket_metrics}). YOLO11s is the sole exception at mAP@0.5-0.95, where XS achieves a marginally higher score than S (0.525~vs.~0.508). Performance is weakest for XXS and M targets, with the latter exhibiting the highest FNR in most configurations. For XXS targets, YOLO11 achieves consistently higher mAP@0.25 than the evaluated YOLOv8 models.

\begin{figure*}[t!]
    \centering
    \includegraphics[width=\textwidth, trim={0cm 16.9cm 0cm 0cm}, clip]{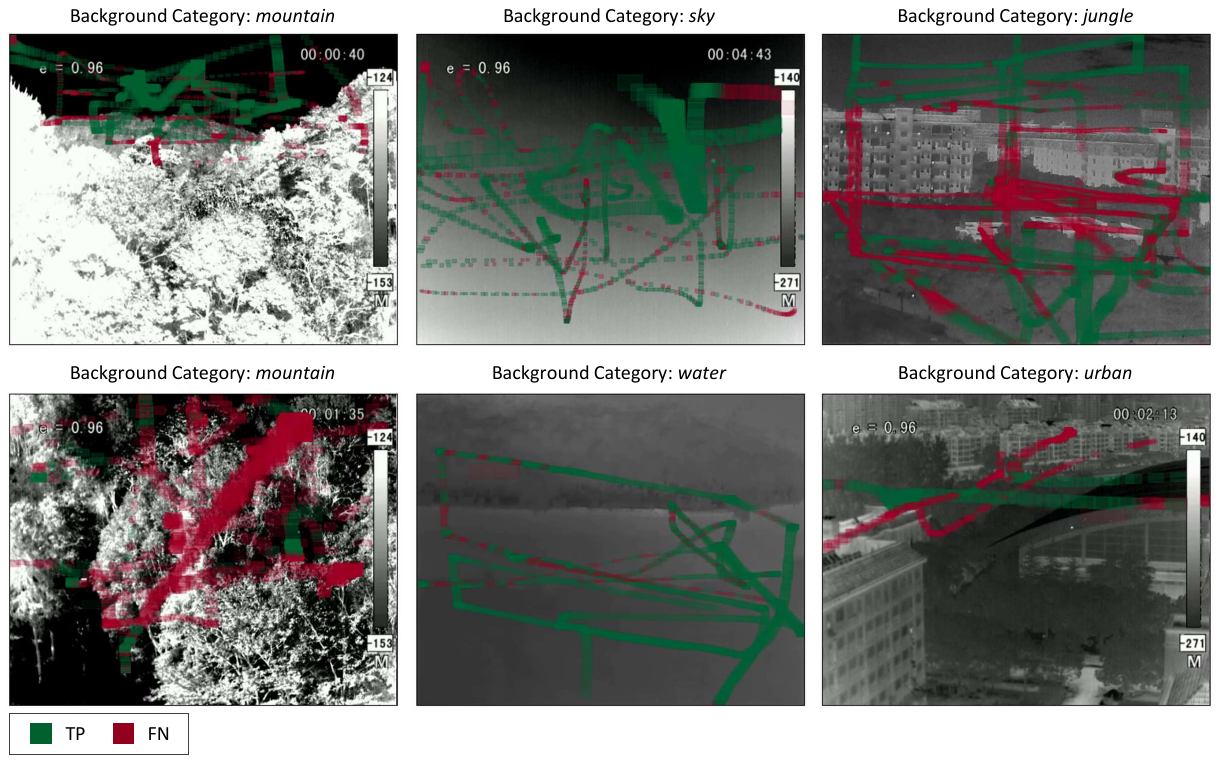}
    \caption{\label{fig:results_heatmaps_TP_FN}Spatial distribution of YOLOv8l detection outcomes across selected CST test sequences~\cite{Xie:2025} spanning diverse background categories. GT boxes are accumulated across frames and superimposed on a representative frame, with true positives (TP) shown in green and false negatives (FN) in red. Detection performance is higher against homogeneous \textit{sky} and \textit{water} backgrounds, whereas FNs are concentrated in areas with vegetation and complex background structures, indicating considerable variability even within the same background category (\textit{mountain}).\vspace{0.3cm}}
\end{figure*}

\vspace{-0.3cm}
\paragraph{Background Category.} Background-dependent performance exhibits a similar pattern: \textit{water} and \textit{sky} yield the highest mAP at less restrictive IoU thresholds and the lowest FNR, whereas \textit{building} exhibits the lowest mAP@0.25 and highest FDR (Tab.~\ref{tab:results_background_cats_all_models}). At the stricter mAP@0.5-0.95 criterion, \textit{sky} occasionally falls below \textit{jungle}, consistent with the localization degradation observed for YOLOv8l. Fig.~\ref{fig:results_heatmaps_TP_FN} further illustrates the spatial distribution of detection outcomes. Missed detections cluster against vegetation and structured backgrounds, whereas targets against open sky and water are detected more reliably, even within the same category.

\begin{table}[t!]
    \centering\footnotesize
    \caption{Detection performance by object-size category across model configurations on the CST test split~\cite{Xie:2025}. Size categories follow the definitions in Tab.~\ref{tab:size_categories}. Per-metric bests are shown in \textbf{bold}, while second-best results are \underline{underlined}. \vspace{-0.2cm}}.
    \label{tab:model_bucket_metrics}

    \begin{tabular}{|c|l|ccc|c|}
        \hline
        YOLO &Size & \multicolumn{3}{c|}{mAP~$\uparrow$} & FNR~$\downarrow$ \\
        Model & Cat. & @0.25 & @0.5 & @0.5-0.95 & \\
        \hline
        \multirow[c]{5}{*}{v8n} & XXS & 0.553 & 0.505 & 0.257 & 0.671 \\
         & XS & \underline{0.728} & \underline{0.720} & \underline{0.468} & \underline{0.463} \\
         & S & \textbf{0.801} & \textbf{0.789} & \textbf{0.495} & \textbf{0.407} \\
         & M & 0.461 & 0.411 & 0.260 & 0.785 \\
         & L & -- & -- & -- & -- \\\hline
         
        \multirow[c]{5}{*}{v8s} & XXS & 0.471 & 0.417 & 0.211 & 0.588 \\
         & XS & \underline{0.752} & \underline{0.737} & \underline{0.448} & \underline{0.333} \\
         & S & \textbf{0.843} & \textbf{0.828} & \textbf{0.529} & \textbf{0.228} \\
         & M & 0.538 & 0.472 & 0.304 & 0.584 \\
         & L & -- & -- & -- & -- \\\hline
         
        \multirow[c]{5}{*}{v8m} & XXS & 0.621 & 0.569 & 0.290 & 0.664 \\
         & XS & \underline{0.776} & \underline{0.771} & \underline{0.502} & \underline{0.444} \\
         & S & \textbf{0.836} & \textbf{0.829} & \textbf{0.523} & \textbf{0.381} \\
         & M & 0.431 & 0.402 & 0.265 & 0.806 \\
         & L & -- & -- & -- & -- \\\hline
         
        \multirow[c]{5}{*}{11s} & XXS & 0.663 & 0.602 & 0.302 & 0.610 \\
         & XS & \underline{0.811} & \underline{0.802} & \textbf{0.525} & \underline{0.392} \\ 
         & S & \textbf{0.835} & \textbf{0.815} & \underline{0.508} & \textbf{0.354} \\
         & M & 0.355 & 0.349 & 0.247 & 0.751 \\
         & L & -- & -- & -- & -- \\\hline
         
        \multirow[c]{5}{*}{11m} & XXS & 0.675 & 0.609 & 0.309 & 0.628 \\
         & XS & \underline{0.793} & \underline{0.786} & \underline{0.512} & \underline{0.412} \\
         & S & \textbf{0.807} & \textbf{0.800} & \textbf{0.517} & \textbf{0.408} \\
         & M & 0.339 & 0.339 & 0.262 & 0.855 \\
         & L & -- & -- & -- & -- \\\hline
         
        \multirow[c]{5}{*}{11l} & XXS & 0.689 & 0.620 & 0.313 & 0.635 \\
         & XS & \underline{0.820} & \underline{0.810} & \underline{0.526} & \underline{0.382} \\
         & S & \textbf{0.863} & \textbf{0.844} & \textbf{0.559} & \textbf{0.301} \\
         & M & 0.522 & 0.504 & 0.366 & 0.721 \\
         & L & -- & -- & -- & -- \\\hline
    \end{tabular}
\end{table}

\begin{table}[t!]
    \centering\footnotesize
    \caption{\label{tab:results_per_model_attrib_metrics}Attribute-specific detection performance across YOLOv8 (n/s/m) and YOLO11 (s/m/l) configurations on the CST test split~\cite{Xie:2025}, stratified by the presence (\ding{51}) or absence (\ding{55}) of annotated instance attributes. For each attribute and metric, the superior result is highlighted in \textbf{bold}.\vspace{-0.2cm}}
    
    \begin{tabular}{|c|l|c|ccc|c|}
    \hline
    YOLO & Attr. & Incl. & \multicolumn{3}{c|}{mAP~$\uparrow$} &  FNR~$\downarrow$ \\
    Model & & & @0.25 & @0.5 & @0.5-0.95 &\\
    \hline
        \multirow[c]{6}{*}{v8n} & \multirow[c]{2}{*}{TC} & \ding{55} & \textbf{0.864} & \textbf{0.843} & \textbf{0.519} & \textbf{0.314} \\
        &  & \ding{51} & 0.542 & 0.517 & 0.307 & 0.700 \\
        \cline{2-7}
        & \multirow[c]{2}{*}{DBC} & \ding{55} & \textbf{0.674} & \textbf{0.655} & \textbf{0.402} & \textbf{0.536} \\
        &  & \ding{51} & 0.653 & 0.596 & 0.323 & 0.562 \\
        \cline{2-7}
        & \multirow[c]{2}{*}{CDB} & \ding{55} & \textbf{0.712} & \textbf{0.689} & \textbf{0.418} & \textbf{0.498} \\
        &  & \ding{51} & 0.497 & 0.480 & 0.298 & 0.707 \\
        \hline
        \multirow[c]{6}{*}{v8s} & \multirow[c]{2}{*}{TC} & \ding{55} & \textbf{0.820} & \textbf{0.797} & \textbf{0.483} & \textbf{0.229} \\
        &  & \ding{51} & 0.555 & 0.519 & 0.290 & 0.557 \\
        \cline{2-7}
        & \multirow[c]{2}{*}{DBC} & \ding{55} & \textbf{0.668} & \textbf{0.639} & \textbf{0.376} & \textbf{0.415} \\
        &  & \ding{51} & 0.618 & 0.553 & 0.293 & 0.476 \\
        \cline{2-7}
        & \multirow[c]{2}{*}{CDB} & \ding{55} & \textbf{0.694} & \textbf{0.662} & \textbf{0.388} & \textbf{0.391} \\
        &  & \ding{51} & 0.532 & 0.512 & 0.296 & 0.538 \\
        \hline
        \multirow[c]{6}{*}{v8m} & \multirow[c]{2}{*}{TC} & \ding{55} & \textbf{0.890} & \textbf{0.867} & \textbf{0.526} & \textbf{0.319} \\
        &  & \ding{51} & 0.610 & 0.591 & 0.363 & 0.671 \\
        \cline{2-7}
        & \multirow[c]{2}{*}{DBC} & \ding{55} & \textbf{0.729} & \textbf{0.711} & \textbf{0.436} & \textbf{0.520} \\ 
        &  & \ding{51} & 0.712 & 0.660 & 0.367 & 0.560 \\
        \cline{2-7}
        & \multirow[c]{2}{*}{CDB} & \ding{55} & \textbf{0.754} & \textbf{0.733} & \textbf{0.445} & \textbf{0.486} \\
        &  & \ding{51} & 0.609 & 0.593 & 0.372 & 0.680 \\
        \hline
        \multirow[c]{6}{*}{11s} & \multirow[c]{2}{*}{TC} & \ding{55} & \textbf{0.929} & \textbf{0.906} & \textbf{0.562} & \textbf{0.253} \\
        &  & \ding{51} & 0.618 & 0.589 & 0.348 & 0.631 \\
        \cline{2-7}
        & \multirow[c]{2}{*}{DBC} & \ding{55} & \textbf{0.761} & \textbf{0.739} & \textbf{0.451} & \textbf{0.470} \\ 
        &  & \ding{51} & 0.742 & 0.671 & 0.374 & 0.505 \\
        \cline{2-7}
        & \multirow[c]{2}{*}{CDB} & \ding{55} & \textbf{0.782} & \textbf{0.757} & \textbf{0.462} & \textbf{0.441} \\
        &  & \ding{51} & 0.661 & 0.630 & 0.375 & 0.602 \\
        \hline
        \multirow[c]{6}{*}{11m} & \multirow[c]{2}{*}{TC} & \ding{55} & \textbf{0.929} & \textbf{0.899} & \textbf{0.551} & \textbf{0.277} \\
        &  & \ding{51} & 0.606 & 0.584 & 0.354 & 0.655 \\
        \cline{2-7}
        & \multirow[c]{2}{*}{DBC} & \ding{55} & \textbf{0.752} & \textbf{0.730} & \textbf{0.450} & \textbf{0.493} \\
        &  & \ding{51} & 0.738 & 0.658 & 0.356 & 0.535 \\
        \cline{2-7}
        & \multirow[c]{2}{*}{CDB} & \ding{55} & \textbf{0.778} & \textbf{0.752} & \textbf{0.460} & \textbf{0.465} \\
        &  & \ding{51} & 0.629 & 0.600 & 0.365 & 0.627 \\
        \hline
        \multirow[c]{6}{*}{11l} & \multirow[c]{2}{*}{TC} & \ding{55} & \textbf{0.944} & \textbf{0.917} & \textbf{0.571} & \textbf{0.254} \\
        &  & \ding{51} & 0.633 & 0.603 & 0.361 & 0.628 \\
        \cline{2-7}
        & \multirow[c]{2}{*}{DBC} & \ding{55} & \textbf{0.779} & \textbf{0.755} & \textbf{0.465} & \textbf{0.469} \\
        &  & \ding{51} & 0.753 & 0.670 & 0.380 & 0.487 \\
        \cline{2-7}
        & \multirow[c]{2}{*}{CDB} & \ding{55} & \textbf{0.803} & \textbf{0.774} & \textbf{0.475} & \textbf{0.438} \\
        &  & \ding{51} & 0.660 & 0.630 & 0.385 & 0.608 \\
        \hline
    \end{tabular}
\end{table}

\begin{figure*}[t!]
    \centering
    \includegraphics[width=\textwidth, trim={0.3cm 23.7cm 0.3cm 0cm}, clip]{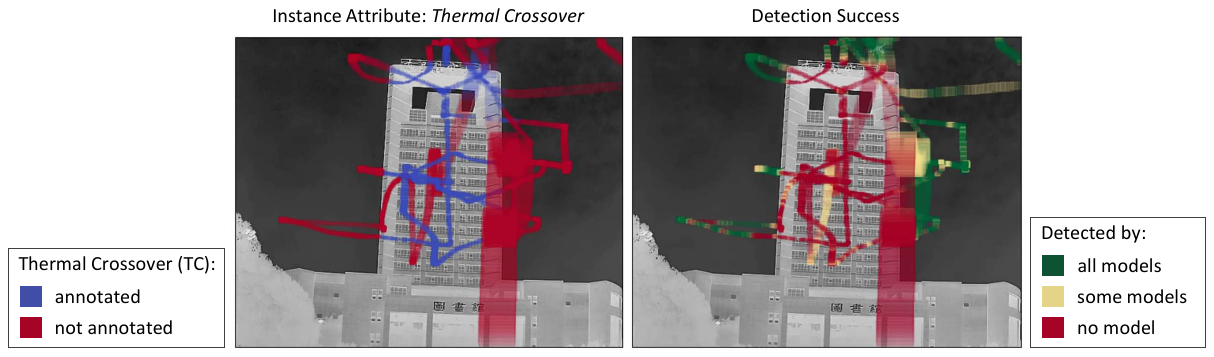}
    \caption{\label{fig:comparison_TC_vs_detection_outcome}Accumulated GT boxes of all building instances in the CST test split~\cite{Xie:2025}, superimposed on a representative frame. \textbf{Left:} instances with (blue) and without (red) TC annotation. \textbf{Right:} detection outcomes across all detectors in Tab.~\ref{tab:model_results_all_metrics}, distinguishing instances detected by all (green), some (yellow), or none (red). TC-annotated instances largely coincide with regions of frequent missed detections along the building facade. However, misses also occur in regions without TC annotation, suggesting a potential role of background structure beyond the annotated attributes and possible limitations in capturing low-contrast conditions through TC annotations.\vspace{0.3cm}}
\end{figure*}

\begin{table*}[t!]
    \centering\footnotesize
    \caption{\label{tab:results_background_cats_all_models}Detection performance by sequence-level background category across model configurations on the CST test split~\cite{Xie:2025}. For each model and metric, the best and second-best results are highlighted in \textbf{bold} and \underline{underlined}, respectively.\vspace{-0.2cm}}
    \begin{tabular}{|c|l|ccc|c|c|}
        \hline
        YOLO & Background & \multicolumn{3}{c|}{mAP~$\uparrow$} & FNR~$\downarrow$ & FDR~$\downarrow$\\
        Model & Category & @0.25 & @0.5 & @0.5-@0.95 & & \\
        \hline
        \multirow[c]{6}{*}{v8n} 
         & water & \textbf{1.000} & \textbf{0.987} & \textbf{0.674} & \textbf{0.089} & \textbf{0.021} \\
         & sky & \underline{0.987} & \underline{0.963} & \underline{0.458} & \underline{0.235} & \underline{0.037} \\
         & jungle & 0.710 & 0.700 & 0.441 & 0.454 & 0.151 \\
         & mountains & 0.667 & 0.630 & 0.416 & 0.522 & 0.039 \\
         & urban & 0.511 & 0.499 & 0.294 & 0.762 & 0.042 \\
         & building & 0.427 & 0.419 & 0.245 & 0.645 & 0.187 \\\hline
         
        \multirow[c]{6}{*}{v8s} 
        & water & \textbf{0.999} & \textbf{0.991} & \textbf{0.675} & \textbf{0.094} & \textbf{0.017} \\
         & sky & \underline{0.979} & \underline{0.942} & \underline{0.441} & \underline{0.183} & \underline{0.056} \\
         & jungle & 0.687 & 0.681 & 0.429 & 0.312 & 0.590 \\
         & mountains & 0.680 & 0.630 & 0.401 & 0.435 & 0.192 \\
         & urban & 0.572 & 0.542 & 0.294 & 0.560 & 0.141 \\
         & building & 0.360 & 0.348 & 0.185 & 0.616 & 0.596 \\\hline
         
        \multirow[c]{6}{*}{v8m} 
        & water & \textbf{0.999} & \textbf{0.976} & \textbf{0.693} & \textbf{0.132} & \textbf{0.015} \\
         & sky & \underline{0.977} & \underline{0.937} & 0.441 & \underline{0.262} & 0.042 \\
         & jungle & 0.782 & 0.771 & \underline{0.482} & 0.457 & \underline{0.017} \\ 
         & mountains & 0.704 & 0.677 & 0.458 & 0.524 & 0.021 \\
         & urban & 0.589 & 0.579 & 0.350 & 0.698 & 0.066 \\
         & building & 0.492 & 0.479 & 0.263 & 0.637 & 0.269 \\\hline
         
        \multirow[c]{6}{*}{11s} 
         & water & \textbf{1.000} & \textbf{0.980} & \textbf{0.687} & \textbf{0.096} & \textbf{0.020} \\
         & sky & \underline{0.987} & \underline{0.958} & 0.445 & \underline{0.236} & 0.038 \\
         & jungle & 0.889 & 0.870 & \underline{0.540} & 0.325 & \underline{0.021} \\
         & mountains & 0.720 & 0.680 & 0.460 & 0.511 & 0.024 \\
         & urban & 0.584 & 0.563 & 0.321 & 0.653 & 0.049 \\
         & building & 0.430 & 0.426 & 0.257 & 0.638 & 0.336 \\\hline
         
        \multirow[c]{6}{*}{11m} & water & \textbf{1.000} & \textbf{0.983} & \textbf{0.700} & \textbf{0.101} & \textbf{0.014} \\
         & sky & \underline{0.990} & \underline{0.950} & 0.451 & \underline{0.274} & 0.045 \\
         & jungle & 0.841 & 0.822 & \underline{0.507} & 0.362 & 0.037 \\
         & mountains & 0.709 & 0.679 & 0.453 & 0.496 & \underline{0.025} \\
         & urban & 0.568 & 0.555 & 0.331 & 0.699 & 0.049 \\
         & building & 0.515 & 0.508 & 0.296 & 0.629 & 0.273 \\\hline
         
        \multirow[c]{6}{*}{11l} & water & \textbf{0.997} & \textbf{0.982} & \textbf{0.686} & \textbf{0.101} & \underline{0.017} \\
         & sky & \underline{0.987} & \underline{0.958} & 0.458 & \underline{0.209} & 0.044 \\
         & jungle & 0.857 & 0.836 & \underline{0.524} & 0.381 & \textbf{0.009} \\
         & mountains & 0.718 & 0.681 & 0.461 & 0.460 & 0.078 \\
         & urban & 0.607 & 0.584 & 0.342 & 0.658 & 0.030 \\
         & building & 0.565 & 0.556 & 0.323 & 0.595 & 0.158 \\
        \hline
    \end{tabular}
\end{table*}

\vspace{-0.3cm}
\paragraph{Instance Attributes.} The presence of annotated instance attributes is consistently associated with lower mAP across IoU thresholds and elevated FNR (Tab.~\ref{tab:results_per_model_attrib_metrics}). The relative effects identified for YOLOv8l extend across models, with thermal crossover (TC) exhibiting the largest FNR increase (0.33-0.39), followed by CDB (0.15-0.21) and DBC (0.02-0.06). Fig.~\ref{fig:comparison_TC_vs_detection_outcome} further illustrates the relationship between TC annotations and detection failures across model configurations in \textit{building} sequences. Instances missed by all detectors predominantly cluster along the building facade and coincide with TC-annotated regions, indicating a consistent failure pattern across models. Nevertheless, such failures also occur in the absence of TC annotations, suggesting that the binary TC labels do not fully characterize the low-contrast conditions associated with the building facade.

{
    \small
    \bibliographystyle{ieeenat_fullname}
    \bibliography{main}
}

\end{document}